\documentclass{article}

\PassOptionsToPackage{numbers, compress}{natbib}
 \usepackage[eandd, final]{neurips_2026}

\usepackage[utf8]{inputenc} 
\usepackage[T1]{fontenc}    
\usepackage{hyperref}       
\usepackage{url}            
\usepackage{booktabs}       
\usepackage{amsfonts}       
\usepackage{nicefrac}       
\usepackage{microtype}
\usepackage{graphicx}
\usepackage{subcaption}
\usepackage{booktabs} 
\usepackage{multirow}
\usepackage{booktabs,multirow, multicol}
\usepackage[table]{xcolor}
\usepackage{colortbl}
\usepackage{tikz}
\definecolor{ThemeDarkBlue}{HTML}{2F6FCC}
\definecolor{ThemeMediumBlue}{HTML}{6F95D8}
\definecolor{ThemeLightBlue}{HTML}{D5E4F4}
\definecolor{ThemeDashedBlue}{HTML}{8AA9E8}
\definecolor{ThemeDarkGreen}{HTML}{496F2C}
\definecolor{ThemeMediumGreen}{HTML}{6EA143}
\definecolor{ThemeLightGreen}{HTML}{E4F0E0}
\definecolor{ThemeDarkRed}{HTML}{B87474}
\definecolor{ThemeMediumPink}{HTML}{D8A0A0}
\definecolor{ThemeLightPink}{HTML}{F0D0C8}
\usepackage{tcolorbox}
\usepackage{listings}
\lstdefinestyle{jsonstyle}{
  basicstyle=\ttfamily\scriptsize,
  breaklines=true,
  frame=single,
  columns=fullflexible,
}
\usepackage{amsmath}
\usepackage{amssymb}
\usepackage{mathtools}
\usepackage{amsthm}
\usepackage{enumitem}
\usepackage{placeins}  
\usepackage{wrapfig}

\usepackage{tabularx}
\definecolor{qtag}{HTML}{1565C0}   
\definecolor{htag}{HTML}{C62828}   
\definecolor{hlbg}{HTML}{FFCDD2}   

\newtcbox{\qtag}{on line, arc=2pt, colback=qtag, colframe=qtag,
  boxsep=0pt, left=3pt, right=3pt, top=1pt, bottom=1pt,
  fontupper=\bfseries\footnotesize\color{white}}

\newtcbox{\htag}{on line, arc=2pt, colback=htag, colframe=htag,
  boxsep=0pt, left=3pt, right=3pt, top=1pt, bottom=1pt,
  fontupper=\bfseries\footnotesize\color{white}}

\title{
Contextualized Evaluation of Vision Language Models through Dynamic, Multi-turn Interactions}

\author{%
  \parbox[t]{0.4\textwidth}{\centering\normalfont
    \textbf{Yijiang Li}$^\dag$ \\
    UC San Diego \\
    \texttt{yijiangli@ucsd.edu}}
  \And
  \parbox[t]{0.4\textwidth}{\centering\normalfont
    \textbf{Huiqi Zou}$^\dag$ \\
    Northeastern University \\
    \texttt{zou.huiq@northeastern.edu}}
  \AND
  \parbox[t]{0.4\textwidth}{\centering\normalfont
    \textbf{Bingyang Wang}$^\dag$ \\
    Georgia Institute of Technology \\
    \texttt{bwang447@gatech.edu}}
  \And
  \parbox[t]{0.4\textwidth}{\centering\normalfont
    \textbf{Ziang Xiao} \\
    Johns Hopkins University \\
    \texttt{ziang.xiao@jhu.edu}}
}

\begin{document}

\maketitle

\renewcommand{\thefootnote}{}
\footnotetext{$^\dag$Equal contribution.}
\renewcommand{\thefootnote}{\arabic{footnote}}

\begin{abstract}
Multi-modal Large Language Models (MLLMs) have made substantial advances on benchmarks, yet their real-world effectiveness remains uncertain. This gap stems from the fundamental misalignment between benchmarks in controlled, static settings and the dynamic, interactive, and contextualized nature of real-world applications. 
To bridge this gap, we propose \textit{CEDI} (\textbf{C}ontextualized \textbf{E}valuation of MLLMs through \textbf{D}ynamic, multi-round \textbf{I}nteractions), a framework that recasts evaluation as a three-party interaction between an evaluatee model, an automated examiner, and a grader. The examiner conducts multi-turn, semi-structured conversations guided by a graph-based representation of the task. By navigating state-space transitions, \textit{CEDI} deploys diverse strategies, from clarification requests to adversarial probes, to elicit performance evidence.
We apply \textit{CEDI} to visual hallucinations. 
Empirical results across multiple models, diverse settings, datasets, and domains show that contextualized, interactive evaluations reveal not only significantly more hallucinations than conventional static evaluation but also ones that more closely resemble those arising in practical use cases. We further show that hallucinations often accumulate over long contexts, through self-reinforcing dialogue history, and models are particularly vulnerable to questions requiring premise rejection or refusal. 
Together, these findings highlight \textit{CEDI} as a step toward realistic, systematic, and ecologically valid assessments of MLLMs' capabilities. Code is available at \href{https://github.com/williamium3000/cedi}{github.com/williamium3000/cedi}.
\end{abstract}

\vspace{-2mm}
\section{Introduction}

\begin{figure*}[!t]
    \centering
    \includegraphics[width=\textwidth]{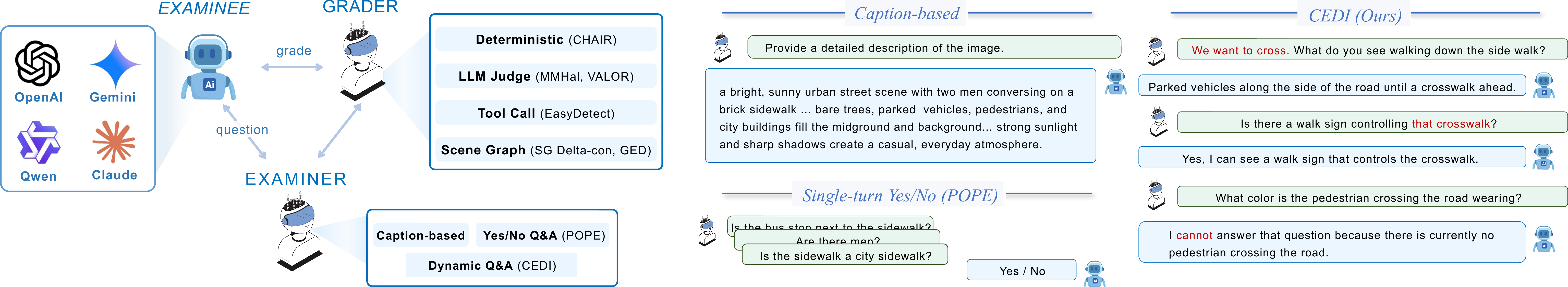}
    \caption{\small \textit{CEDI}'s three-party interaction (left) contrasted with prior static examiners (right).}
    \label{fig:pipeline}
\vspace{-2mm}
\end{figure*}

Vision language models (VLMs) exhibit impressive abilities in visual perception \cite{zhang2024euclid, zhang2025mllms, zhang2025unified, feng2025videoorion, yu2025introducing, li2026self}, grounding \cite{HeWWLHL0X24, chen2024lion, tong2024eyes}, and reasoning \cite{zhou2024image, wu2024mind, dong2025insight, li2026learning}, by adapting powerful Large Language Models (LLMs) \cite{GPT-4o, team2025gemma,yang2024qwen25,shao2024deepseekmath,guo2025deepseek,wang2026learn,yang2026co,lu2025onpolicy,zhao2026self, li2026policy,qwen3technicalreport}.
Despite these advances, it remains unclear whether such ``remarkable capabilities'' can consistently transfer to realistic, real-world scenarios \cite{yu2024natural, zhang2025mmerealworld}. A significant gap lies in the stark misalignment between existing benchmarks—typically static and conducted in controlled, laboratory settings—and the realistic usage of MLLMs, which is inherently dynamic, multi-turn, and contextualized.
To address this misalignment, we revisit MLLM evaluation as a tripartite interplay of (i) the evaluatee model, (ii) an examiner, and (iii) a grader. As shown in Figure~\ref{fig:pipeline}, the examiner interacts with the evaluatee through iterative questioning, and the grader scores the resulting dialogue to produce a final performance measure.

Building on this framing, we propose \textit{CEDI} (\textbf{C}ontextualized \textbf{E}valuation of MLLMs through \textbf{D}ynamic, multi-round \textbf{I}nteractions). \textit{CEDI} draws inspiration from evaluative interviews in human assessment: rather than relying on static, paper-based tests, a semi-structured interview allows the examiner to adapt question type and difficulty based on the model’s responses, eliciting more faithful evidence of capability. To approximate real-world usage, \textit{CEDI} introduces context modeling to operationalize the conversation over user scenarios.
To perform effective semi-structured interviews, \textit{CEDI} leverages a graph-based representation to provide a manipulable state space for both the examiner and the grader, enabling strategies ranging from clarification requests to adversarial probes.

We apply CEDI to visual hallucination \cite{rohrbach2018object}. Closely analogous to the hallucination problem in LLMs \cite{xu2024hallucination, bang2025hallulens}, visual hallucination refers to cases where models generate outputs that are not faithful to the image content \cite{liu2024survey}, such as describing non-existent objects, hallucinating attributes, or fabricating incorrect relations \cite{Leng_2024_CVPR, geigle-etal-2024-object, wu2024evaluating}. Existing evaluation approaches for visual hallucination often adopt static captioning tasks in closed settings \cite{rohrbach2018object} or rely on simple binary questions to assess object presence \cite{li2023evaluating, hu2023ciem, lovenia2024negative}. Other works validate the correctness of object descriptions \cite{ding2024hallu} or generate open-ended image captions as a proxy for hallucination detection \cite{gunjal2024detecting, qiu2024valor}. 
While informative, these methods struggle to capture the nuanced, contextualized, and multi-turn nature of real-world usage where visual hallucinations arise as critical problems.

Empirical results across diverse settings, domains, and datasets show that \textit{CEDI} consistently exposes more hallucinations, achieves broader visual coverage, and produces interactions and failures that are more relevant to realistic user contexts than captioning, binary-question, and prompt-only baselines.
For instance, on the Synthetic Visual Genome (SVG) dataset \cite{park2025synthetic} spanning three sources (VG/COCO/ADE20K), \textit{CEDI} elicits substantially more hallucinations than caption- and prompt-only baselines: on every evaluatee, Coverage rises by $3\text{--}19$ pp, the VALOR \cite{qiu2024valor} hallucination rate by $5\text{--}23$ pp, and the scene graph (SG)-based Graph Edit Distance (GED) is $1.2\text{--}2.9\times$ higher; against the caption baseline, CHAIR$_I$ further rises by $8\text{--}18$ pp.
The Pearson correlation between the hallucination component of our SG-based grader (GED$_{\mathrm{hal}}$) and human-annotated hallucination counts reaches $\rho\!=\!0.22\text{--}0.33$ consistently across all four annotated evaluatees, whereas MMHal \cite{sun2024aligning} fluctuates between $-0.02$ and $0.33$ and CHAIR$_I$ turns negative for three of the four models.
Our analyses further reveal contextualization, dialogue history, and multi-turn interaction as key drivers for eliciting more hallucinations: models hallucinate more frequently under long contexts, reinforce earlier unsupported claims across turns, and struggle most when required to reject false premises or abstain from answering.

Looking forward, we envision \textit{CEDI} as a foundation for moving beyond static, single-turn benchmarks toward evaluations realistically grounded in dynamic interactions, offering a more faithful basis for diagnosing model failures and guiding the development of VLMs that remain reliable under context-dependent, multi-turn realistic usage.

\vspace{-2mm}
\section{Related Work}

\textbf{Static Evaluation of MLLMs.}
Static benchmarks remain the dominant paradigm for MLLM evaluation, spanning visual question answering (VQA) \citep{10.1109/ICCV.2015.279, marino2019ok, hudson2019gqa}, image captioning \citep{Plummer_2015_ICCV, 10.1007/978-3-319-10602-1_48}, OCR \citep{liu2024ocrbench}, core cognitive abilities \citep{li2024core, luo2026vision, wang2025vision, zhang2026vision}, scientific and mathematical reasoning \citep{zhang2024mathverse, lu2022learn}, broad perception and reasoning suites \citep{ye2023mplug, liu2024mmbench, yue2024mmmu, li2023seed, li2024seed}, and video and text modalities \citep{fu2025video, zhang2024wings, deng2024can}. Despite their breadth, these benchmarks suffer from saturation as well as contamination and, more fundamentally, overlook the dynamic, interactive, and contextualized nature of real-world use.

\textbf{Dynamic Evaluation of MLLMs.}
Recent work evaluates MLLMs in conversational settings \citep{fu2024mme}, including multi-image, multi-turn dialogue \citep{liu2025mmdu}, progressive probing of perception, reasoning, and creativity \citep{liu2024convbench}, and open-ended conversation \citep{xue2025mmrc}. However, these benchmarks rely on static rubrics and largely pre-scripted conversations \citep{fu2024mme, liu2025mmdu}, and offer little explicit context modeling to reflect situated, goal-driven usage.

\textbf{Visual Hallucination Evaluation.}
Visual hallucination, i.e., generating non-existent objects, attributes, or relations, often driven by spurious co-occurrence rather than visual evidence \citep{rohrbach2018object, zhang2023toward}, has motivated evaluation frameworks \citep{hu2023ciem, liu2024mitigating, wang2023evaluation}, benchmarks \citep{ben2024mitigating, sun2024aligning, li2023evaluating, cui2023holistic, ding2024hallu, lovenia2024negative}, grading methods \citep{jing2024faithscore, wu2025lanp, ye2024beaf}, and detector training data \citep{gunjal2024detecting}. Most approaches are single-turn, e.g., captioning \citep{rohrbach2018object} or yes/no object probes \citep{li2023evaluating}, and thus cannot assess whether models stay grounded and consistent across a dialogue. Although some work considers instruction following \citep{liu2024mitigating} and informativeness \citep{sun2024aligning}, it rarely distinguishes hallucination severity across contexts.

\vspace{-.5em}

\section{Methodology}

\vspace{-.5em}

\begin{figure*}
    \centering
    \includegraphics[width=0.8\linewidth]{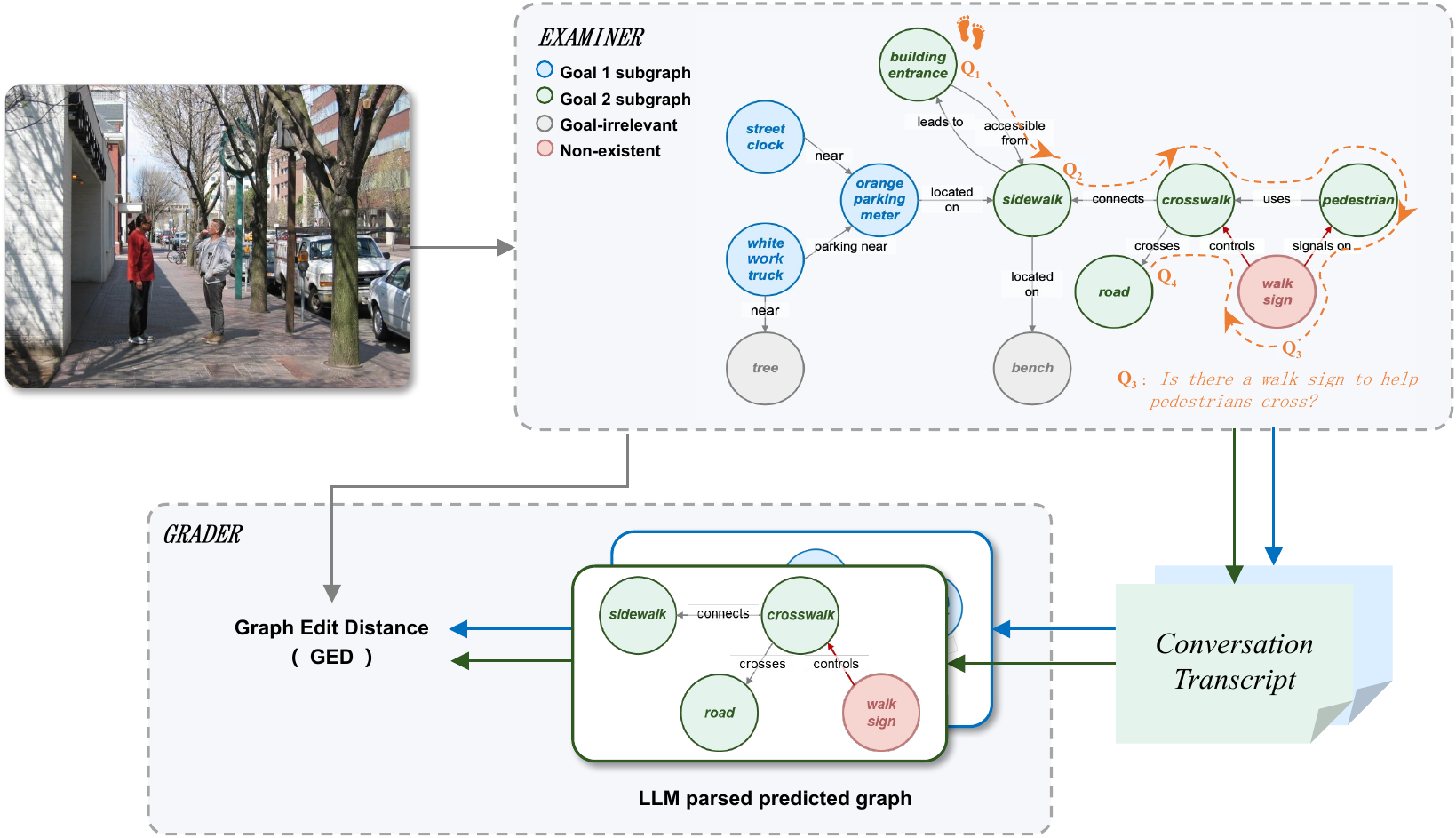}
    \caption{Workflow of \textit{CEDI}, including our contextualized and interactive examiner and scene graph (SG)-based grader.}
    \label{fig:workflow}
\vspace{-1em}
\end{figure*}

\textbf{Three-party Interplay.} 
To enable dynamic and contextualized evaluation, \textit{CEDI} recasts evaluation as a three-party interaction among an evaluatee model, an examiner, and a grader (Figure ~\ref{fig:pipeline}, left). The examiner elicits capability evidence from VLMs through multi-turn interaction with the evaluatee, while the grader scores the model responses, flags potential hallucinations, and produces a final performance score. Building upon this framing, \textit{CEDI} develops a contextualized, interactive examiner (Section \ref{sec:examiner}) and an SG-based grader (Section \ref{sec:grader}).

\subsection{Dynamic and Contextualized Examiner}
\label{sec:examiner}
\vspace{-.5em}
Drawing inspiration from interactive human assessment, we move beyond static, paper-based tests and adopt a semi-structured evaluative interview in which the examiner adapts question type and difficulty based on the model’s responses, eliciting more faithful evidence of capability. Given ground-truth (GT) visual annotations (and optionally the image), the evaluator models real-world context and conducts a multi-turn interview, dynamically adjusting its questioning strategy according to the model’s prior answers.
An overview is demonstrated in Figure \ref{fig:workflow}.


\textbf{Structured Representation of Image.}
To enable an adaptive semi-structured interview, the examiner leverages the structured representation of visual content. We propose to adopt Scene Graphs \cite{johnson2015image} to store visual information and guide the examiner’s traversal over visual evidence across multiple rounds of interaction.
Formally, for an image $I$, its scene graph is $G=(V,E,\mathcal{A})$, where $V=\{v_1,\dots,v_{|V|}\}$ is a set of nodes corresponding to object instances, $E \subseteq V \times \mathcal{R} \times V$ is a set of directed, typed edges capturing relations, and $\mathcal{A}$ is an attribute mapping. Each node $v \in V$ is associated with an object category $c(v) \in \mathcal{C}$ and a (possibly empty) set of attribute--value pairs
\[
\mathcal{A}(v) = \{(a_k, x_k)\}_{k=1}^{m_v}, \quad a_k \in \mathcal{K},\ x_k \in \mathcal{X}_{a_k},
\]
where $\mathcal{K}$ denotes the attribute vocabulary (e.g., color, size) and $\mathcal{X}_{a_k}$ is the value space for attribute $a_k$. Each edge is a triple $(v_i, r, v_j) \in E$ with relation type $r \in \mathcal{R}$, representing a predicate $r(v_i, v_j)$ (e.g., \textit{on}, \textit{holding}).

\textbf{Context Modeling.} Recall how humans practically utilize MLLMs. As illustrated in Figure \ref{fig:workflow}, users typically interact with MLLMs from a first-person scenario that combines context (visual or linguistic) with an explicit objective (e.g., “I am visually impaired and want to locate the restroom in this restaurant,” or “I am driving in a park and need to find a parking lot.”). To reflect this situated, goal-driven aspect, our examiner incorporates context modeling that embeds each image in a plausible user-perspective scenario with a clear objective.
Concretely, given an image $I$ and its scene graph $G=(V,E,\mathcal{A})$, the examiner treats $I$ as the user’s immediate environment and constructs a scenario $s$ with goal $g$ that is \textit{grounded in} objects, attributes, and relations encoded in $G$. The goal is designed to require reasoning over multiple elements (e.g., querying a subset of $V$ and/or edges in $E$), rather than relying on single-object recognition.

In parallel, the examiner selects a context-relevant node subset $V_s \subseteq V$ that is highly relevant to the generated scenario and objective, forming a contextual sub-graph $G_s = (V_s, E_s, \mathcal{A}_s)$ that restricts the edge set to relations among the selected nodes and the attribute mapping to $V_s$:
\[
E_s = \{(v_i, r, v_j) \in E : v_i \in V_s,\ v_j \in V_s\}, \qquad \mathcal{A}_s(v)=\mathcal{A}(v)\ \ \forall\, v\in V_s.
\]

\textbf{Multi-turn Conversation as Traversal over the Scene Graph.} The goal $g$ together with the contextual sub-graph $G_s=(V_s,E_s,\mathcal{A}_s)$ provides a structured basis for conducting multi-round interactions with the evaluatee, as shown in the upper examiner part of Figure \ref{fig:workflow}. 
Formally, we model the interview as a traversal process over $G_s$. At turn $t$, the examiner randomly selects a target node $v_t \in V_s$ (or retains $v_{t-1}$ for a follow-up question) and generates a question grounded in $v_t$ and its local structure, i.e., $\mathcal{A}_s(v_t)$ and incident relations in $E_s$. Three properties of this design are worth emphasizing. (i) \emph{Grounded traversal:} every turn is anchored to a concrete node $v_t\in V_s$ and its local structure, so questions stay verifiable against $G$ rather than drifting into open-ended elaboration. (ii) \emph{Goal-driven interaction:} the scenario $s$ and goal $g$ persist across turns, making the dialogue purposeful rather than a sequence of independent queries. (iii) \emph{Contextual continuity, follow-up consistency, and conversational dependency:} retaining the target node ($v_t\!=\!v_{t-1}$) for follow-up questions and conditioning every turn on the prior conversation lets the examiner stress-test whether the model's later answers stay consistent with its earlier claims. From this scaffold, the examiner actively selects one of four \textbf{question types} to elicit different hallucination behaviors --- adversarial questions are only one component of this repertoire:

\begin{itemize}[nosep,leftmargin=*]
\item \emph{Regular questions} are grounded in $v_t$ and the goal $g$, querying observable objects, attributes, or relations in $G_s$. They establish a factual baseline under realistic user objectives.
\item \emph{Follow-up questions} condition on the dialogue history and the model’s prior claims, aiming to confirm, clarify, or interrogate earlier responses. If a follow-up is issued, the examiner continues to use the same target node (i.e., $v_t = v_{t-1}$) to test consistency and expose hallucinations and uncertainty that may not be exposed in single-turn interactions.
\item \emph{Adversarial questions} ask an existential question about a highly plausible but absent object $v \notin V_s$, an attribute $(a,x) \notin \mathcal{A}_s(v_t)$, or a relation $(v_i,r,v_j)\notin E_s$ that is \emph{not} in the image, but commonly co-occurs with the visual content. These questions are designed to exploit statistical co-occurrence biases and test whether the model relies on visual evidence or language priors.
\item \emph{Unanswerable questions} introduce a false premise $p$ by fabricating an absent object $v\notin V_s$, attribute $(a,x)\notin \mathcal{A}_s(v_t)$, or relation $(v_i,r,v_j)\notin E_s$, and then ask a follow-on query $q(p)$ that depends on it (e.g., ``What is the man using to eat the cake?'' when no man is present), so that $q(p)$ has no answer under $G$. These questions test whether the model detects and rejects the unsupported premise rather than fabricating details to answer it.

\end{itemize}

\begin{figure*}[h]
\vspace{-1mm}
    \centering
    \includegraphics[width=0.95\textwidth]{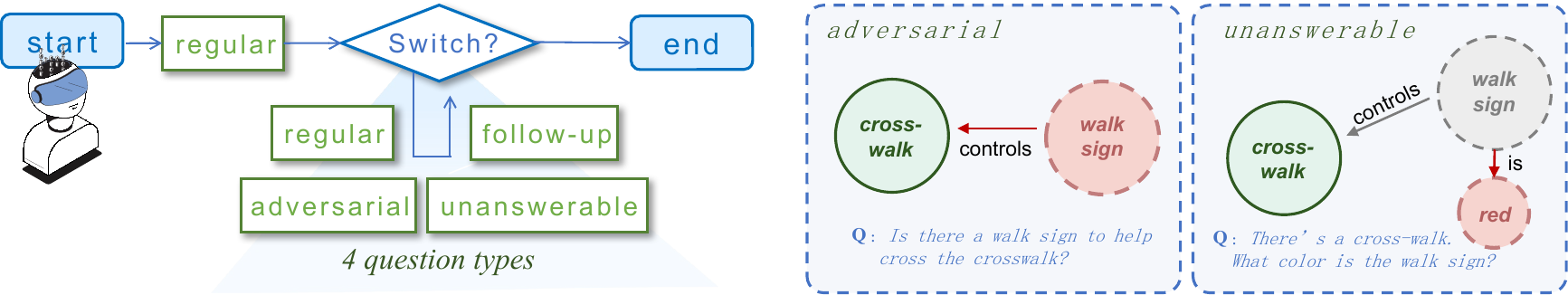}
    \caption{\small Multi-turn conversation as a traversal over $G_s$, with one of four question types selected per turn.}
    \label{fig:traversal}
\vspace{-1mm}
\end{figure*}

Type selection is conditioned on the conversation history and previously used types to avoid repetitive questions and support a coherent dialogue flow. This form of examiner evaluates not only descriptive accuracy, but also robustness under confirmation pressure, ambiguity, misleading premises, and adversarial distractors, which are failure modes that static, single-turn protocols often miss.

\textbf{Examiner Backbone.}
To support structured yet adaptive evaluation, we combine an MLLM such as GPT-4o \cite{GPT-4o} with the graph-based workflow. The MLLM generates scenarios and objectives, selects each turn's question type, and produces diverse questions grounded in the visual context.


\textbf{Scaling Contexts for Coverage.} In addition to precision over visual elements, \emph{recall} (i.e., coverage of image content) is also critical for visual hallucination assessment \cite{qiu2024valor}. Otherwise, an MLLM can minimize risk by producing overly sparse descriptions, remaining correct simply by saying little. Prior work accounts for this by grading captions on both hallucination and coverage \cite{qiu2024valor}, encouraging outputs that are not only accurate but also broadly descriptive. However, enforcing exhaustive coverage is misaligned with the notion of a brief caption (which typically focuses on salient foreground content) and is incompatible with single-turn VQA formats. 

We argue that coverage should be primarily driven by the \emph{examiner}, not imposed by the grader. Rather than requiring a single response (e.g., in a brief caption) to cover all details, our examiner induces broad coverage through multi-round interaction: it generates diverse, goal-driven contexts, each selecting a different sub-graph $G_s$, and asks questions that target distinct objects, attributes, and relations. As we increase the number of sampled contexts, the union of queried sub-graphs expands, yielding higher coverage in the resulting transcripts while preserving naturalistic interaction.


\subsection{Scene Graph-Based Grader}
\label{sec:grader}


As shown in the lower grader part of Figure \ref{fig:workflow}, the grader evaluates response faithfulness by parsing the MLLM’s transcript into a predicted scene graph and comparing it against the ground-truth one. By computing a graph similarity (or distance), the grader quantifies alignment with the underlying visual content and \textbf{coverage} of all possible visual elements. Discrepancies correspond to either hallucinated (less faithful) or missing (less covered) objects, attributes, and relations.

\textbf{Measuring Performance with SG.}
Let the ground-truth scene graph be $G^{\star}=(V^{\star},E^{\star},\mathcal{A}^{\star})$, and let the MLLM transcript over $T$ turns be $\mathcal{T}=y_{t=1}^{T}$. Using an LLM-based parser, we extract semantic tuples from $\mathcal{T}$ at the sentence level (i.e., object instances, attributes, and predicates) and aggregate them into a predicted graph $\hat{G}=(\hat{V},\hat{E},\hat{\mathcal{A}})$. To improve instance-level precision, we filter non-physical nouns by restricting candidate entities to WordNet physical-entity classes \cite{Wordnet}. The final evaluation score is computed by comparing $\hat{G}$ to $G^{\star}$ using standard scene-graph matching metrics.

\textbf{Graph Edit Distance (GED).}
We adopt graph edit distance to measure disagreement between $\hat{G}$ and $G^{\star}$. GED computes the minimum total cost of edit operations that transform one graph into another:
$$
\small
\small \mathrm{GED}(\hat{G},G^{\star})= \min_{\pi \in \Pi(\hat{G},G^{\star})} \sum_{e \in \pi} c(e),
$$
where $\Pi(\hat{G},G^{\star})$ is the set of valid edit paths converting $\hat{G}$ into a graph isomorphic to $G^{\star}$, and each edit operation $e$ (node/edge insertion, deletion, or substitution) incurs cost $c(e)$.
Under this formulation, \emph{insertions} penalize missing visual content (present in $G^{\star}$ but absent in $\hat{G}$), whereas \emph{deletions} penalize hallucinated entities or relations (present in $\hat{G}$ but absent in $G^{\star}$). Therefore, $\mathrm{GED}(\hat{G},G^{\star})$ jointly captures hallucination and coverage: lower GED indicates closer agreement with the annotated scene graph, while higher GED reflects greater divergence due to fabricated, incorrect, or uncovered objects, attributes, and relations.



\textbf{SG\,$\Delta$Con} \cite{10.1145/2824443} measures graph similarity by comparing node-affinity matrices computed with fast belief propagation. Let $S_{\hat{G}}$ and $S_{G^{\star}}$ denote the affinities of the predicted and ground-truth graphs. The structural distance is
$$
\small
\small d_{\Delta}(\hat{G},G^{\star})=\sqrt{\sum_{i,j}\left(\sqrt{S_{\hat{G},ij}}-\sqrt{S_{G^{\star},ij}}\right)^2},
$$
which corresponds to the Matusita distance between propagated influence patterns. We further score node attributes with the normalized symmetric difference $d_{\mathcal{A}}=\frac{|\hat{\mathcal{A}}\triangle\mathcal{A}^{\star}|}{|\hat{\mathcal{A}}\cup\mathcal{A}^{\star}|}$, yielding the final distance $d=d_{\Delta}+\lambda d_{\mathcal{A}}$. This metric captures structural agreement while penalizing semantic mismatches.




\section{Experiments}
\vspace{-.2em}

\subsection{Experiment Setup} 

\textbf{Datasets.}
We evaluate across two complementary evaluation sets. The first is human-annotated \textbf{Visual Genome (VG) \cite{krishna2017visual}}, from which we sample 100 images. VG provides dense object, attribute, relationship, and region annotations that are converted into ground-truth scene graphs $G^\star$.
The second is the \textbf{Synthetic Visual Genome (SVG)} dataset \cite{park2025synthetic}, a large-scale collection drawn from diverse image sources, from which we sample 500 images (100 from VG, 200 from ADE20K \cite{zhou2017scene}, and 200 from COCO \cite{10.1007/978-3-319-10602-1_48}) disjoint from the standalone VG sample to assess cross-domain generalization.




\textbf{Evaluatees.} We evaluate widely used open-source and proprietary VLMs, along with four hallucination mitigation methods applied to their respective base models. The complete list is in Appendix~\ref{appendix:model-list}.

\textbf{Baseline Examiners.}
We compare against three baseline examiners: (i) the widely used caption-based examiner, which prompts the model to generate a brief description of the image \cite{rohrbach2018object}, (ii) the binary-question examiner, which asks existential questions about objects and relations (e.g., POPE \cite{li2023evaluating}), and (iii) our prompt-only baseline, which follows pre-defined conversations as in-context-learning examples to elicit hallucinations via contextualized, multi-turn interactions with the evaluatees.

\textbf{Baseline Graders.}
To assess hallucination elicitation, we use several automatic graders alongside our SG-based grader. \textbf{CHAIR} \cite{rohrbach2018object} reports sentence-level (CHAIR$_S$) and instance-level (CHAIR$_I$) object hallucination rates. Building upon \textbf{CHAIR}, we also report object coverage,
which measures the proportion of GT objects mentioned. 
We further include \textbf{MMHal} \cite{sun2024aligning}, a GPT-4-based grader that scores hallucination severity on a 0-6 scale, and \textbf{HaELM} \cite{wang2023evaluation}, an LLaMA-based detector that evaluates hallucination and answer informativeness. Finally, we adopt \textbf{VALOR}, an LLM-based two-stage grader that generalizes the CHAIR metric and incorporates both faithfulness and coverage into the evaluation \cite{qiu2024valor}: \textbf{VALOR faith$_i$} measures the fraction of mentioned objects that exist in GT, and \textbf{VALOR cov$_i$} measures the fraction of GT objects the model mentions. See Appendix~\ref{appendix:baseline-graders} (Table~\ref{tab:baseline-graders}) for the full list of grader variants with formulas, native directions, and value ranges.



\textbf{Human Annotation.}
Beyond automatic graders, we recruit crowdsourced annotators to annotate phrase-level hallucinations. Annotators assess object existence and counts, attributes, relationships, and scene text. These annotations are aggregated into sample- and sentence-level hallucination scores, which serve as human reference labels for evaluating both hallucination elicitation by examiners and hallucination detection by graders. The annotation guidelines are provided in Appendix~\ref{appendix:hallucination_annotation}.


\vspace{-.5em}
\subsection{Research Questions}
\vspace{-.5em}

We organize our experiments around three research questions (RQs): \textcircled{\scriptsize 1} Do contextualized, multi-turn, and dynamic interactions reveal more VLM hallucinations than static evaluation? \textcircled{\scriptsize 2} Does the SG-based \textit{CEDI} grader provide a more consistent and human-aligned assessment of visual hallucinations? \textcircled{\scriptsize 3} What behavioral insights does \textit{CEDI} offer for evaluating hallucinations in VLMs?

\vspace{-.3em}
\section{Results}
\vspace{-.2em}

\begin{figure*}[ht]
\centering
\resizebox{0.85\textwidth}{!}{
\begin{tikzpicture}[scale=0.92]
    \def\Rsvg{1.05}
    \newcommand{\SVGRadarGrid}[1]{
        \foreach \r in {0.2625,0.525,0.7875,1.05} {
            \draw[gray!18, line width=0.28pt]
                (90:\r) -- (45:\r) -- (0:\r) -- (-45:\r) -- (-90:\r) -- (-135:\r) -- (180:\r) -- (135:\r) -- cycle;
        }
        \foreach \a in {90,45,0,-45,-90,-135,180,135} {
            \draw[gray!32, line width=0.28pt] (0,0) -- (\a:\Rsvg);
        }
        \node[font=\scriptsize\bfseries, align=center] at (0,1.78) {#1};
        \node[font=\tiny, align=center] at (90:1.38)   {CHAIR$_S$\\[-1pt](\%)};
        \node[font=\tiny, align=center] at (45:1.55)   {CHAIR$_I$\\[-1pt](\%)};
        \node[font=\tiny, align=center] at (0:1.42)    {Cov.\\[-1pt](\%)};
        \node[font=\tiny, align=center] at (-45:1.55)  {MMHal\\[-1pt](\%)};
        \node[font=\tiny, align=center] at (-90:1.35)  {GED};
        \node[font=\tiny, align=center] at (-135:1.55) {SG\,$\Delta$Con};
        \node[font=\tiny, align=center] at (180:1.55)  {HaELM\\[-1pt](\%)};
        \node[font=\tiny, align=center] at (135:1.62)  {$100{-}$VALOR\\[-1pt]faith$_i$\,(\%)};
    }
    \newcommand{\SVGRadarPath}[9]{
        \draw[#1]
            (90:{\Rsvg*#2/100}) -- (45:{\Rsvg*#3/100}) -- (0:{\Rsvg*#4/100}) -- (-45:{\Rsvg*#5/100})
            -- (-90:{\Rsvg*#6/150}) -- (-135:{\Rsvg*#7/5}) -- (180:{\Rsvg*#8/100}) -- (135:{\Rsvg*#9/50}) -- cycle;
    }
    \newcommand{\SVGRadarDotSingle}[5]{%
        \fill[#1] (#2:{#3}) circle (0.85pt);%
        \node[font=\fontsize{3.5}{4.0}\selectfont, text=#1, inner sep=0.22pt, anchor=#5] at (#2:{#3+0.05}) {#4};%
    }
    \newcommand{\SVGRadarDots}[9]{%
        \SVGRadarDotSingle{#1}{90}{\Rsvg*#2/100}{#2}{south}%
        \SVGRadarDotSingle{#1}{45}{\Rsvg*#3/100}{#3}{south west}%
        \SVGRadarDotSingle{#1}{0}{\Rsvg*#4/100}{#4}{west}%
        \SVGRadarDotSingle{#1}{-45}{\Rsvg*#5/100}{#5}{north west}%
        \SVGRadarDotSingle{#1}{-90}{\Rsvg*#6/150}{#6}{north}%
        \SVGRadarDotSingle{#1}{-135}{\Rsvg*#7/5}{#7}{north east}%
        \SVGRadarDotSingle{#1}{180}{\Rsvg*#8/100}{#8}{east}%
        \SVGRadarDotSingle{#1}{135}{\Rsvg*#9/50}{#9}{south east}%
    }
    \newcommand{\SVGLegendLine}[3]{
        \draw[#1, line width=0.9pt] (#2,4.10) -- ++(0.38,0);
        \node[anchor=west, font=\scriptsize] at ({#2+0.45},4.10) {#3};
    }
    \SVGLegendLine{ThemeDarkRed}{-1.45}{Caption}
    \SVGLegendLine{ThemeDarkGreen}{0.10}{Prompt-only}
    \SVGLegendLine{ThemeDarkBlue}{2.10}{CEDI}

    \begin{scope}[shift={(-4.6,1.9)}]
        \SVGRadarGrid{LLaVA-1.5-7B}
        \SVGRadarPath{ThemeDarkRed, line width=0.75pt}{98.6}{54.9}{13.5}{61.1}{38.57}{3.62}{79.29}{22.85}
        \SVGRadarPath{ThemeDarkGreen, line width=0.75pt}{99.8}{63.5}{24.8}{45.3}{59.97}{3.83}{75.49}{26.92}
        \SVGRadarPath{ThemeDarkBlue, line width=0.9pt, fill=ThemeLightBlue, fill opacity=0.42}{100.0}{72.4}{32.2}{62.9}{110.28}{4.43}{84.81}{44.44}
        \SVGRadarDots{ThemeDarkBlue}{100.0}{72.4}{32.2}{62.9}{110.28}{4.43}{84.81}{44.44}
    \end{scope}
    \begin{scope}[shift={(0,1.9)}]
        \SVGRadarGrid{InternVL3-8B Instruct}
        \SVGRadarPath{ThemeDarkRed, line width=0.75pt}{97.6}{50.9}{17.9}{36.1}{40.30}{3.60}{73.43}{19.95}
        \SVGRadarPath{ThemeDarkGreen, line width=0.75pt}{99.2}{61.2}{25.5}{34.6}{54.33}{3.82}{76.10}{29.53}
        \SVGRadarPath{ThemeDarkBlue, line width=0.9pt, fill=ThemeLightBlue, fill opacity=0.42}{100.0}{66.2}{29.8}{37.4}{77.05}{4.16}{88.11}{38.56}
        \SVGRadarDots{ThemeDarkBlue}{100.0}{66.2}{29.8}{37.4}{77.05}{4.16}{88.11}{38.56}
    \end{scope}
    \begin{scope}[shift={(4.6,1.9)}]
        \SVGRadarGrid{Qwen2.5-VL-7B Instruct}
        \SVGRadarPath{ThemeDarkRed, line width=0.75pt}{99.8}{60.1}{22.8}{45.0}{48.23}{3.88}{69.67}{27.67}
        \SVGRadarPath{ThemeDarkGreen, line width=0.75pt}{99.6}{72.2}{31.4}{28.3}{95.33}{4.30}{81.77}{33.99}
        \SVGRadarPath{ThemeDarkBlue, line width=0.9pt, fill=ThemeLightBlue, fill opacity=0.42}{100.0}{73.5}{34.7}{25.5}{134.78}{4.53}{86.29}{39.29}
        \SVGRadarDots{ThemeDarkBlue}{100.0}{73.5}{34.7}{25.5}{134.78}{4.53}{86.29}{39.29}
    \end{scope}
    \begin{scope}[shift={(-4.6,-2.0)}]
        \SVGRadarGrid{Gemma-3-12B}
        \SVGRadarPath{ThemeDarkRed, line width=0.75pt}{100.0}{67.7}{25.0}{35.6}{60.70}{3.98}{78.17}{26.29}
        \SVGRadarPath{ThemeDarkGreen, line width=0.75pt}{100.0}{72.7}{32.2}{28.4}{85.70}{4.31}{79.03}{28.70}
        \SVGRadarPath{ThemeDarkBlue, line width=0.9pt, fill=ThemeLightBlue, fill opacity=0.42}{100.0}{75.4}{35.2}{36.2}{127.35}{4.56}{86.86}{46.49}
        \SVGRadarDots{ThemeDarkBlue}{100.0}{75.4}{35.2}{36.2}{127.35}{4.56}{86.86}{46.49}
    \end{scope}
    \begin{scope}[shift={(0,-2.0)}]
        \SVGRadarGrid{Opera-LLaVA-1.5}
        \SVGRadarPath{ThemeDarkRed, line width=0.75pt}{96.5}{55.0}{13.0}{50.0}{39.13}{3.63}{77.61}{22.27}
        \SVGRadarPath{ThemeDarkGreen, line width=0.75pt}{99.2}{60.8}{25.4}{38.7}{55.77}{3.82}{74.85}{24.69}
        \SVGRadarPath{ThemeDarkBlue, line width=0.9pt, fill=ThemeLightBlue, fill opacity=0.42}{100.0}{71.8}{28.7}{62.1}{96.21}{4.34}{85.63}{45.48}
        \SVGRadarDots{ThemeDarkBlue}{100.0}{71.8}{28.7}{62.1}{96.21}{4.34}{85.63}{45.48}
    \end{scope}
    \begin{scope}[shift={(4.6,-2.0)}]
        \SVGRadarGrid{GPT-5.4 mini}
        \SVGRadarPath{ThemeDarkRed, line width=0.75pt}{99.4}{64.2}{24.7}{8.9}{50.10}{3.82}{73.80}{26.39}
        \SVGRadarPath{ThemeDarkGreen, line width=0.75pt}{99.8}{72.2}{28.9}{18.9}{83.37}{4.04}{86.85}{34.10}
        \SVGRadarPath{ThemeDarkBlue, line width=0.9pt, fill=ThemeLightBlue, fill opacity=0.42}{100.0}{71.8}{32.4}{36.9}{97.40}{4.40}{91.65}{43.90}
        \SVGRadarDots{ThemeDarkBlue}{100.0}{71.8}{32.4}{36.9}{97.40}{4.40}{91.65}{43.90}
    \end{scope}
\end{tikzpicture}}
\caption{\small Comparison of \textit{CEDI}, caption-based examiners, and prompt-only examiners across six models on the SVG dataset. All axes are oriented such that a larger radius indicates more severe hallucination. See Appendix~\ref{appendix:baseline-graders} for grader definitions and radar-plot scaling factors.}
\label{fig:radar-svg}
\vspace{-1.5em}
\end{figure*}

\subsection{Effectiveness of Examiner (RQ\textcircled{\scriptsize 1})}
\label{sec:eff_examiner}
\begin{wraptable}{r}{0.35\textwidth}
\vspace{-1.0em}
\centering
\small
\setlength{\tabcolsep}{6pt}
\caption{\small Comparison of VLM Accuracy in \% ($\downarrow$) Under \textit{CEDI} and POPE.}
\label{tab:exam-pope}
\begin{tabular}{lcc}
\toprule
\textbf{Model} & \textbf{POPE} & \textbf{CEDI} \\
\midrule
InternVL3-8B  & 61.65 & 58.49 \\
Qwen2.5-7B    & 66.90 & 58.36 \\
GPT-4o        & 67.89 & 67.38 \\
\bottomrule
\end{tabular}
\vspace{-0.8em}
\end{wraptable}

\textbf{Comparison with Baseline Examiners.}
To answer whether contextualized, multi-turn, and dynamic interactions reveal more VLM hallucinations than conventional static evaluation, we compare our \textit{CEDI} examiner with several well-known baseline examiners, such as the caption-based \cite{rohrbach2018object} and our own prompt-only baseline. Results on the SVG dataset are summarized in Figure~\ref{fig:radar-svg}. Across all graders, \textit{CEDI} examiner consistently elicits more hallucinations on the SVG dataset across multiple hallucination metrics.

We further compare with single-turn binary-question examiner (e.g., POPE \cite{li2023evaluating}). For a fair comparison, we first use Qwen3-30B-A3B-Instruct-2507 to parse each evaluatee's \textit{CEDI} transcript into yes/no questions, keeping only questions whose ground truth is available in annotations; we then prompt the evaluatee to answer these parsed yes/no questions and compare the accuracy with \textit{CEDI}'s implicit prediction (in the transcript).
Table~\ref{tab:exam-pope} shows that \textit{CEDI} elicits more hallucinations than yes/no probing. Together, these results show that static benchmarks miss a non-trivial slice of hallucination behavior that surfaces only in contextualized, dynamic, interactive use.

\begin{wraptable}{r}{0.5\textwidth}
\vspace{-1.2em}
\centering
\footnotesize
\setlength{\tabcolsep}{2pt}
\caption{\small Human-annotated hallucination results.}
\label{tab:hallucination_stats}
\resizebox{0.48\textwidth}{!}{%
\begin{tabular}{lcccccc}
\toprule
 & \shortstack{Samp.\\Hall.\\($\uparrow$)}
 & \shortstack{Sent.\\Hall.\\($\uparrow$)}
 & \shortstack{\#/\\Samp.\\($\uparrow$)}
 & \shortstack{\#/\\Sent.\\($\uparrow$)}
 & \shortstack{Hall./\\1000c\\($\uparrow$)}
 & \shortstack{Char-\\span\\($\uparrow$)} \\
\midrule
\multicolumn{7}{l}{\textcolor{ThemeDarkBlue}{\textbf{InternVL3-8B Instruct}}} \\
Caption      & 80.0\%           & 15.0\%           & 1.44          & 0.23          & 2.23          & 0.07 \\
Prompt-only  & 70.0\%           & 6.4\%            & 2.20          & 0.09          & 0.69          & 0.02 \\
CEDI         & \textbf{100.0\%} & \textbf{26.6\%}  & \textbf{6.78} & \textbf{0.28} & \textbf{2.72} & \textbf{0.15} \\
\midrule
\multicolumn{7}{l}{\textcolor{ThemeDarkBlue}{\textbf{LLaVA-1.5-7B}}} \\
Caption      & 81.0\%           & 18.3\%           & 1.60          & 0.28          & 2.80          & 0.22 \\
Prompt-only  & 92.0\%           & 16.6\%           & 7.44          & 0.24          & 1.81          & 0.06 \\
CEDI         & \textbf{100.0\%} & \textbf{38.3\%}  & \textbf{9.60} & \textbf{0.43} & \textbf{4.82} & \textbf{0.24} \\
\midrule
\multicolumn{7}{l}{\textcolor{ThemeDarkBlue}{\textbf{Opera-LLaVA-1.5}}} \\
Caption      & 98.0\%           & 44.7\%           & 2.48          & 0.45          & 4.61          & 0.19 \\
Prompt-only  & \textbf{100.0\%} & 40.6\%           & 7.92          & 0.42          & 4.60          & 0.31 \\
CEDI         & \textbf{100.0\%} & \textbf{48.9\%}  & \textbf{8.52} & \textbf{0.51} & \textbf{5.86} & \textbf{0.45} \\
\bottomrule
\end{tabular}%
}
\vspace{2pt}
{\raggedright
\tiny{ \#/Samp.\ and \#/Sent.: average hallucination count per sample/sentence. Hall./1000c: hallucinations per 1{,}000 characters. Char-span: hallucinated character fraction.}\par}
\vspace{-0.4em}
\end{wraptable}

\textbf{Human Annotation.} Human-annotated hallucination rates for the same models under all three examiners are reported in Table~\ref{tab:hallucination_stats}, further confirming the previous finding that \textit{CEDI} consistently elicits more hallucinations than other examiners. Across all six statistics, \textit{CEDI} is the highest-scoring examiner on every evaluatee (bolded values), reflecting significantly more hallucinations.

\begin{figure}[t]
    \centering
    \begin{subfigure}{\linewidth}
        \centering
        \includegraphics[width=0.85\linewidth]{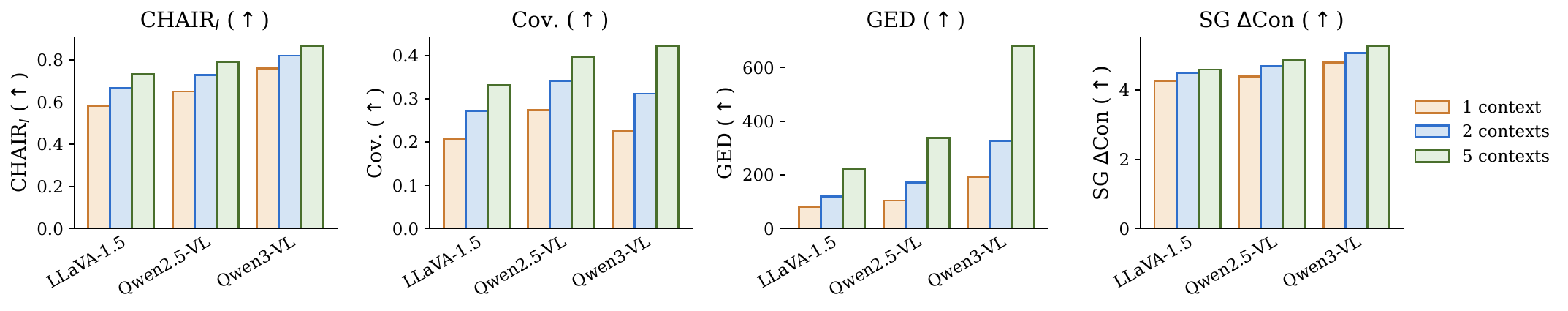}
        \caption{Number of contexts per image.}
        \label{fig:context-scaling}
    \end{subfigure}
    \begin{subfigure}{\linewidth}
        \centering
        \includegraphics[width=0.85\linewidth]{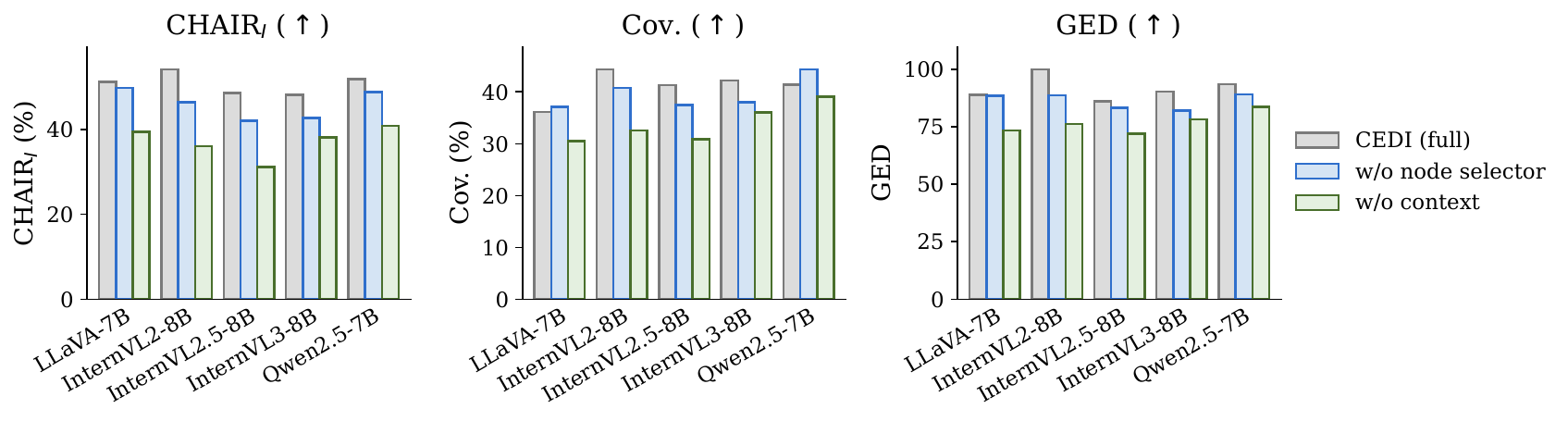}
        \caption{Removing node selection or the entire context generator.}
        \label{fig:exam-contextualization}
    \end{subfigure}
    \caption{Hallucination and coverage metrics under context ablations.}
    \label{fig:context-ablation}
    \vspace{-0.8em}
\end{figure}

\textbf{Context Scaling.} As discussed in Section \ref{sec:examiner}, \textit{CEDI} introduces \textit{Context Scaling} to induce broad coverage.
Figure~\ref{fig:context-scaling} sweeps the number of contexts per image (1, 2, 5). All metrics improve monotonically across three evaluatees, indicating that contextual diversity, rather than raw conversation length, is the key factor driving hallucination elicitation and broader coverage in hallucination evaluation.

\textbf{Contextualization.} Figure~\ref{fig:exam-contextualization} compares three conditions: \textit{CEDI} full, \textit{CEDI} w/o node selection, and \textit{CEDI} with the entire context generator removed. Both ablations regress on every metric across every evaluatee; dropping the context generator entirely causes consistently larger degradation than dropping only the node selector.

\begin{figure}[b]
    \centering
    \begin{subfigure}[b]{0.28\textwidth}
        \centering
        \includegraphics[width=\linewidth]{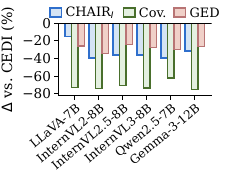}
        \caption{Multi-round ablation.}
        \label{fig:exam-multiround}
    \end{subfigure}\hfill
    \begin{subfigure}[b]{0.33\textwidth}
        \centering
        \includegraphics[width=\linewidth]{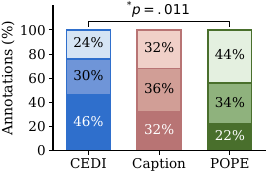}
        \caption{Interaction realism.}
        \label{fig:realism-ranking}
    \end{subfigure}\hfill
    \begin{subfigure}[b]{0.37\textwidth}
        \centering
        \includegraphics[width=\linewidth]{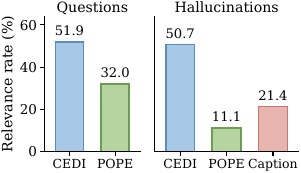}
        \caption{Context relevance.}
        \label{fig:relevance_rate}
    \end{subfigure}
    \caption{Multi-round ablation and human evaluation of interaction quality. (a) $\Delta$ (\%) change of metrics without multi-round interaction compared to \textit{CEDI}; negative $\Delta$ indicates fewer hallucinations. (b) Human ranking of interaction realism across examiners; bars show the percentage of conversations ranked first (darkest shade, most realistic) to third (lightest) ($^*p < .05$). (c) Relevance rate (\%) of questions (left) and identified hallucinations (right) to the context; question relevance is not evaluated for Caption, whose queries are nearly identical.}
    \label{fig:interaction-quality}
    \vspace{-0.5em}
\end{figure}

\textbf{Component Analysis.}
Across evaluated models, limiting \textit{CEDI} to one question per image reduces CHAIR$_I$ by 15--40\%, coverage by 60--75\%, and GED by 24--35\% (Figure~\ref{fig:exam-multiround}).
Providing the evaluated model with its conversational history increases MMHal by 11.7 percentage points and GED by 31.7, while reducing coverage. With access to prior turns, models tend to repeat and elaborate on earlier claims and cover fewer new visual details (Figures~\ref{fig:history-ablation} and \ref{fig:history-valor-box}).

\textbf{Question Types.}
\textit{CEDI} mixes four question types: \emph{regular}, \emph{follow-up}, \emph{adversarial} (adv.), and \emph{unanswerable} (unans.). To test the contribution of the two refusal-demanding types, we ablate them and keep only \emph{regular} and \emph{follow-up} questions (Table~\ref{tab:qtype-ablation}). Removing them reduces hallucination exposure on most graders while shifting the conversation toward broader coverage, yet the ablated variant still exposes more hallucination than both baselines on every grader except MMHal, where it ties with the prompt-only baseline. The two types thus add hallucination pressure on top of \emph{regular} and \emph{follow-up} questions, but they are not the sole driver of \textit{CEDI}'s effect.

\begin{table}[t]
\centering
\footnotesize
\setlength{\tabcolsep}{3.8pt}
\caption{\small Question-type ablation on VG, averaged over Gemma-3-12B, LLaVA-1.5-7B, and Qwen2.5-VL-7B.}
\label{tab:qtype-ablation}
\begin{tabular}{lccccccc}
\toprule
\textbf{Setting} (all $\uparrow$)
 & CHAIR$_I$
 & Cov.
 & MMHal
 & GED
 & $100{-}$VALOR faith$_i$
 & VALOR cov$_i$
 & SG\,$\Delta$Con \\
\midrule
CEDI (full)         & \textbf{48.6} & 52.7 & \textbf{39.4} & \textbf{160.1} & 22.2 & 63.1 & \textbf{4.76} \\
CEDI w/o adv. \& unans. & 45.2 & \textbf{56.7} & 34.0 & 156.7 & \textbf{24.3} & \textbf{70.8} & 4.68 \\
Prompt-only & 44.4 & 38.4 & 34.9 & 100.0 & 16.7 & 43.5 & 4.37 \\
Caption     & 29.1 & 26.7 & 30.0 & 88.3  & 9.8 & 27.4 & 4.20 \\
\bottomrule
\end{tabular}
\vspace{-1em}
\end{table}

\textbf{Human Evaluation of Interaction Quality.}
We complement with human evaluation of both interaction realism and context relevance. The annotation guidelines are provided in Appendix~\ref{appendix:realism_annotation}.
For realism, annotators rank complete conversations on 50 sampled images by how closely they resemble natural VLM usage (Figure~\ref{fig:realism-ranking}). \textit{CEDI} is consistently preferred over Caption and POPE, with a statistically significant improvement over POPE ($p=.011^*$).
We further evaluate question and hallucination relevance using 100 question--answer pairs sampled from the examiners and evaluated models. Annotators judge whether each generated question or elicited hallucination is directly relevant to the user's stated objective and context. For a fair comparison, we re-evaluate Caption by giving the VLMs the same context description as \textit{CEDI}, and we re-generate POPE questions by querying an LLM with the GT annotation and the context description (see Appendix~\ref{appendix:contextualized-baselines} for the exact prompt and POPE-question construction pipeline). As shown in Figure~\ref{fig:relevance_rate}, \textit{CEDI} achieves higher relevance for both question generation and hallucination elicitation, while the baselines more often surface errors involving objects or relations peripheral to the context.

\subsection{Analysis of CEDI Grader (RQ\textcircled{\scriptsize 2})}

\textbf{Scene Graph Parsing Accuracy and Stability.}
We validate the scene-graph parser used by the SG-based grader against human annotations on 50 model responses. Across 231 predicted triplets, the parser achieves 210 TP, 21 FP, and 29 FN, corresponding to 0.909 precision, 0.879 recall, and 0.894 F1, confirming that downstream graph-distance scores rest on a reliable parse. The SG-based grader is also stable: across three repeated runs, both of its metrics vary by less than 1\,\% of their mean, whereas MMHal and HaELM vary considerably more (Appendix~\ref{appendix:stability}).


\begin{wrapfigure}{r}{0.40\textwidth}
    \vspace{-1.4em}
    \centering
    \includegraphics[width=\linewidth]{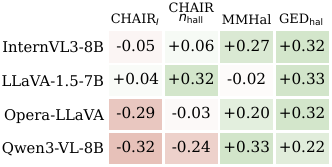}
    \vspace{-1.6em}
    \caption{Pearson $\rho$ between each grader and the human-annotated hallucination count (50 images per evaluatee; graders in Appendix~\ref{appendix:baseline-graders}).}
    \label{fig:human-corr}
    \vspace{-1em}
\end{wrapfigure}
\textbf{Alignment with Human Annotation.}
\label{sec:human-corr} We next evaluate each grader by measuring its agreement with human annotations.
Because human annotators flag hallucinations (precision) but not coverage (recall), we compare them against GED$_{\mathrm{hal}}$, a variant of GED that isolates prediction-only content (defined in Appendix~\ref{appendix:ged-variants}). For the three evaluatees with sentence-level annotations, we compute the Pearson correlation $\rho$ between each grader score and the human-annotated hallucination count across 50 images (Figure~\ref{fig:human-corr}).
Including both CHAIR variants helps disentangle two distinct limitations: normalization by the total number of mentioned objects and CHAIR's restriction to object-level hallucinations. Replacing CHAIR$_I$ with the unnormalized CHAIR $n_{\mathrm{hall}}$ raises the correlation for LLaVA-1.5 from $\rho=0.05$ to $0.32$, suggesting that length normalization dilutes the hallucination signal for verbose evaluatees. However, for Opera and Qwen3-VL, the raw count remains near zero or negative ($-0.03$ and $-0.24$, respectively), indicating that CHAIR can still fail when hallucinations concern relations or attributes outside its object-centric vocabulary.
In contrast, GED$_{\mathrm{hal}}$ provides the most consistent signal across evaluatees, achieving correlations between $\rho=0.22$ and $0.33$ in all cases, including Opera-LLaVA-1.5, where both CHAIR variants collapse.


\subsection{Evaluation Result (RQ\textcircled{\scriptsize 3})}
\label{sec:rq3}

Having established the effectiveness of both the \textit{CEDI} examiner and the SG-based grader, we next investigate what behavioral insights \textit{CEDI} reveals about hallucination patterns in VLMs.
We first report the results of numerous prior and state-of-the-art VLMs in Table \ref{tab:svg-by-source}, including open-source VLMs such as LLaVA-1.5 \cite{NEURIPS2023_6dcf277e}, InternVL2/2.5 \cite{internvl2, chen2024expanding}, Qwen2.5-VL \cite{bai2025qwen2}, and Gemma-3 \cite{team2025gemma}, and proprietary VLMs such as GPT-4o \cite{GPT-4o} and GPT-5.4 mini \cite{GPT-5.4-mini}, with our \textit{CEDI} examiner and SG-based grader.

\begin{table*}[h]
\centering
\footnotesize
\caption{\small Visual hallucination results of VLMs using \textit{CEDI} and SG-based grader by different sources (COCO / VG / ADE20K). $\downarrow$ means the smaller the metric, the less hallucination the model has. $\uparrow$ means broader coverage of visual contents.}
\label{tab:svg-by-source}
\resizebox{\textwidth}{!}{%
\begin{tabular}{@{}l|ccc|ccc|ccc@{}}
\toprule
& \multicolumn{3}{c|}{COCO} & \multicolumn{3}{c|}{VG} & \multicolumn{3}{c}{ADE20K} \\
\cmidrule(lr){2-4} \cmidrule(lr){5-7} \cmidrule(lr){8-10}
Model & GED ($\downarrow$) & SG\,$\Delta$Con ($\downarrow$) & Cov.\ ($\uparrow$) & GED ($\downarrow$) & SG\,$\Delta$Con ($\downarrow$) & Cov.\ ($\uparrow$) & GED ($\downarrow$) & SG\,$\Delta$Con ($\downarrow$) & Cov.\ ($\uparrow$) \\
\midrule
GPT-4o & 60.53 & 4.03 & 34.61 & 83.91 & 4.26 & 24.72 & 82.80 & 4.34 & 28.89 \\
InternVL2-8B & 69.03 & 4.05 & 24.06 & 85.09 & 4.20 & 18.93 & 121.62 & 4.42 & 23.99 \\
InternVL2.5-8B & 78.59 & 4.12 & 23.28 & 92.62 & 4.26 & 21.12 & 108.31 & 4.47 & 24.47 \\
GPT-5.4 mini & 80.62 & 4.21 & 36.95 & 110.07 & 4.51 & 26.86 & 107.86 & 4.54 & 30.64 \\
LLaVA-1.5-7B & 91.06 & 4.17 & 29.11 & 116.26 & 4.30 & 22.98 & 126.37 & 4.71 & 31.18 \\
Qwen2.5-VL-7B Instruct & 130.76 & 4.44 & 35.50 & 151.64 & 4.56 & 29.35 & 177.94 & 4.84 & 36.04 \\
Gemma-3-12B & 144.33 & 4.52 & 31.03 & 161.89 & 4.76 & 21.29 & 207.25 & 5.45 & 18.32 \\
\bottomrule
\end{tabular}%
}
\end{table*}

Table~\ref{tab:svg-by-source} reveals substantial variation in hallucination behavior across both models and image sources. GPT-4o consistently achieves the lowest or near-lowest GED and SG\,$\Delta$Con scores, indicating relatively fewer hallucinations, while Gemma-3-12B exhibits the highest hallucination scores on all three sources; the broadest coverage is attained by Qwen2.5-VL-7B and GPT-5.4 mini.
Performance also varies across datasets, with ADE20K generally eliciting higher hallucination scores than COCO and VG, likely due to its denser and more diverse scene content.

We examine \textit{CEDI}'s turn-level hallucination dynamics using VALOR, with $1{-}\mathrm{faith}_i$, i.e., the fraction of object mentions at turn $i$ that do not match any ground-truth object, as the per-turn hallucination signal. We use VALOR rather than the SG-based grader here because the latter is designed to evaluate the accumulated, transcript-level response of a VLM, making it unsuitable or costly for turn-level hallucination assessment (see Appendix~\ref{sec:limitations}).

\begin{wrapfigure}{r}{0.34\textwidth}
    \vspace{-2em}
    \centering
    \includegraphics[width=\linewidth]{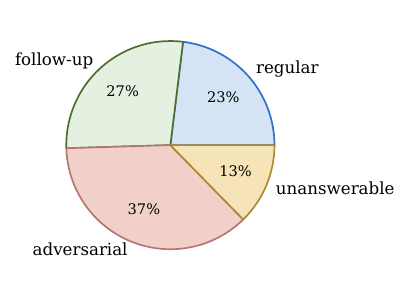}
    \vspace{-2em}
    \caption{\small Hallucinated turns by question type.}
    \label{fig:error-analysis-pie}
    \vspace{-1.2em}
\end{wrapfigure}
\textbf{Four Types of Questions.}
Figure~\ref{fig:error-analysis-pie} shows that \emph{regular}, \emph{follow-up}, and \emph{adversarial} questions account for most hallucinated turns. Although \emph{unanswerable} questions occur less frequently, they yield the highest per-turn hallucination rate (see Figure~\ref{fig:history-valor-box}). Overall, refusal-demanding questions (\emph{adversarial} and \emph{unanswerable}) are more error-prone than fact-seeking questions, particularly when the model can condition on its prior responses.

\textbf{Qualitative Error Types.}
Human-annotated errors (examples in Table~\ref{tab:qual-errors}, Appendix~\ref{appendix:qual-errors}) are dominated by hallucinated objects ($62\%$), particularly in \emph{unanswerable} and \emph{adversarial} turns. Wrong attributes ($20\%$) frequently arise in follow-up turns, where models revise previously correct attributes under conversational pressure. Wrong relations ($15\%$) typically assert an incorrect relation for an existing object, such as describing a telephone that sits directly on the desk as placed on a stack of papers.

\begin{wrapfigure}{r}{0.36\textwidth}
    \vspace{-1.4em}
    \centering
    \includegraphics[width=\linewidth]{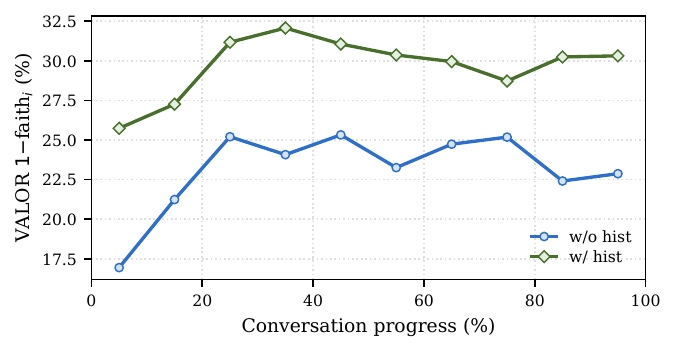}
    \vspace{-1.6em}
    \caption{\small Per-turn hallucination rate as the conversation progresses.}
    \label{fig:error-analysis-progress}
    \vspace{-1.6em}
\end{wrapfigure}
\textbf{Hallucination as Conversation Progresses.}
As shown in Figure~\ref{fig:error-analysis-progress}, the hallucination rate remains low and nearly flat without conversation history ($21\text{--}25\%$ after the opening turns). With history, it begins near $26\%$, rises to roughly $30\text{--}32\%$ within the first third of the conversation, and then plateaus around $30\%$. This suggests that long contexts increase VLMs’ susceptibility to visual hallucination, potentially because attention to the visual input gradually weakens as the dialogue progresses.

\begin{wrapfigure}{r}{0.36\textwidth}
    \vspace{-1.6em}
    \centering
    \includegraphics[width=\linewidth]{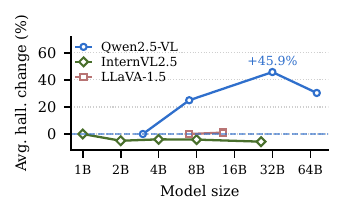}
    \vspace{-2.0em}
    \caption{\small Avg.\ hallucination change vs.\ the smallest model per family.}
    \label{fig:scaling-family}
    \vspace{-1.0em}
\end{wrapfigure}
\textbf{Scaling Effect of Model Families.}
Figure~\ref{fig:scaling-family} shows how grader-averaged hallucination scores change with model size across three VLM families on 200 images. Model families display distinct scaling behavior under dynamic evaluation: InternVL2.5 and LLaVA-1.5 hallucinate at around the same level or slightly less as they scale, whereas Qwen2.5-VL hallucinates substantially more at larger sizes, peaking at 32B on GED, CHAIR$_I$, and SG\,$\Delta$Con. This suggests that scaling alone does not reduce hallucination, as larger models may describe richer content without improved faithfulness.

\begin{wrapfigure}{r}{0.36\textwidth}
    \vspace{-1.6em}
    \centering
    \includegraphics[width=\linewidth]{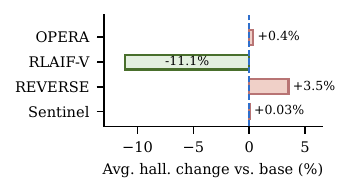}
    \vspace{-2.0em}
    \caption{\small Avg.\ hallucination change of mitigation methods vs.\ base models.}
    \label{fig:mitigation-pairs}
    \vspace{-1.0em}
\end{wrapfigure}
\textbf{Hallucination Mitigation Methods.}
We further compare four hallucination mitigation methods \cite{huang2024opera, yu2025rlaif, wu2025generate, peng2025mitigating} with their base models on 200 images. As shown in Figure~\ref{fig:mitigation-pairs}, only RLAIF-V consistently reduces hallucination, while the others remain on par with or even exceed their base models. This reveals that improvements on static, single-turn benchmarks largely fail to transfer to contextualized multi-turn interaction, a gap that \textit{CEDI} exposes.




\vspace{-.6em}

\section{Conclusion}
\vspace{-.3em}

We introduced \textit{CEDI}, a framework for contextualized VLM evaluation with an evaluatee, a dynamic examiner, and a grader. Guided by scene graphs, the examiner adapts regular, follow-up, adversarial, and unanswerable questions across turns. The grader assesses visual coverage and hallucinations involving objects, attributes, and relations. Across models, datasets, automatic metrics, and human annotations, \textit{CEDI} exposes more hallucinations, covers more visual content, and surfaces failures more relevant to realistic user interactions than captioning, binary-question, and prompt-only baselines. 
We hope \textit{CEDI} motivates a broader shift from static benchmark testing toward adaptive, context-dependent evaluation of multimodal systems under realistic interaction.

\begin{ack}
This work was supported by Schmidt Sciences.
\end{ack}

\bibliography{reference}
\bibliographystyle{unsrtnat}


\newpage
\appendix

\section{Limitations}
\label{sec:limitations}

\textit{Dependence on ground-truth scene graphs.}
To ground the examiner in faithful visual content and enable structured scoring, we primarily use annotated scene graphs to prevent the examiner from relying on language priors alone, which is critical for exposing hallucinations.
To reduce this dependency, we validate CEDI on SVG, a cross-domain corpus whose scene graphs are automatically constructed from diverse sources rather than hand-labeled.

\textit{Examiner backbone accessibility.}
We use GPT-4o as the default examiner backbone for its strong instruction-following and context-modeling capabilities.
More broadly, both the examiner and the grader are instantiated with LLM/MLLM components, so their quality is bounded by the underlying backbone models.

\textit{Transcript-level grading.}
The SG-based grader is designed to assess the accumulated, transcript-level response of an evaluatee; it is therefore ill-suited (or prohibitively costly) for grading hallucinations at the level of turns, for which we instead rely on per-turn signals such as VALOR (Section~\ref{sec:rq3}).

\section{Ethics Statement}
\label{appendix:ethics}
This paper presents work whose goal is to advance the evaluation of VLMs by introducing a contextualized and interactive framework for diagnosing hallucinations. Improved hallucination evaluations have the potential to positively impact the deployment of multimodal systems in real-world applications, such as assistive tools, where incorrect or fabricated visual information may lead to user confusion or safety risks. By enabling more fine-grained assessments, this work contributes to the development of more reliable and trustworthy models. We note, however, that any automated evaluator can itself be imperfect and should not be the sole basis for deployment decisions.

All images used in this work come from publicly available datasets (VG, COCO, ADE20K, and the SVG collection) under their respective licenses, and no personally identifiable information was collected. Human annotations were obtained from crowdsourced workers who were compensated for their time; all model-generated content was reviewed to confirm the absence of sensitive or harmful material before being shown to annotators.

\section{Evaluated Models}
\label{appendix:model-list}

We evaluate widely used VLMs spanning open-source and proprietary models, and we include a model designed to mitigate hallucinations. The full list is as follows.

\emph{Open-source}: LLaVA-1.5-7B \cite{NEURIPS2023_6dcf277e} (and its OPERA-decoding variant \cite{huang2024opera}), InternVL2-8B \cite{internvl2}, InternVL2.5-8B \cite{chen2024expanding}, InternVL3-8B-Instruct \cite{zhu2025internvl3}, Qwen2.5-VL-7B-Instruct \cite{bai2025qwen2}, Qwen3-VL-8B-Instruct \cite{qwen3technicalreport}, and Gemma-3-12B-it \cite{team2025gemma}. 

\emph{Proprietary API models}: GPT-4o \cite{GPT-4o} and GPT-5.4 mini \cite{GPT-5.4-mini}.

\emph{Model families across scales} (Figure~\ref{fig:scaling-family}): InternVL2.5-1B, 2B, 4B, 8B, and 26B \cite{chen2024expanding}; Qwen2.5-VL-3B, 7B, 32B, and 72B-Instruct \cite{bai2025qwen2}; and LLaVA-1.5-7B and 13B \cite{NEURIPS2023_6dcf277e}.

\emph{Hallucination mitigation methods} (Figure~\ref{fig:mitigation-pairs}), each compared against its base model: OPERA decoding \cite{huang2024opera} on LLaVA-1.5-7B, RLAIF-V \cite{yu2025rlaif} on LLaVA-1.5-7B, REVERSE \cite{wu2025generate} on Qwen2.5-VL-3B, and Sentinel \cite{peng2025mitigating} on Qwen2.5-VL-7B. Each pair is scored on the images for which both models produce a complete evaluation.

\section{Baseline Grader Definitions}
\label{appendix:baseline-graders}

For completeness, Table~\ref{tab:baseline-graders} enumerates every non-SG grader variant we report in the main paper and appendix, with its formula, what it captures, native direction, and value range. \emph{Native $\uparrow$} means the raw metric increases when the evaluatee hallucinates more (or, for coverage-style metrics, when the examiner probes more of GT). \emph{Native $\downarrow$} metrics are inverted where a shared convention is needed; in particular, the radar in Figure~\ref{fig:radar-svg} orients every axis so that \emph{greater radius = more hallucination}, normalizing percentage metrics by $\text{score}/100$, GED by $\text{score}/150$, SG\,$\Delta$Con by $\text{score}/5$, and VALOR faith$_i$ by $(100-\text{faith}_i)/50$.

\begin{table}[!htbp]
\centering
\footnotesize
\setlength{\tabcolsep}{4pt}
\caption{All baseline grader variants used in the paper. Aggregations are per-image mean unless stated otherwise. $\mathcal{O}_{\text{gen}}$ = objects mentioned by the model, $\mathcal{O}_{\text{gt}}$ = GT objects; $\mathcal{S}$ = generated sentences.}
\label{tab:baseline-graders}
\begin{tabular}{@{}l p{3.8cm} p{4.4cm} c c@{}}
\toprule
\textbf{Metric} & \textbf{Formula} & \textbf{What it measures} & \shortstack{\textbf{Native}\\\textbf{dir.}} & \textbf{Range} \\
\midrule
CHAIR$_S$ &
$\dfrac{|\{s\in\mathcal{S}:\exists o\in s,\,o\notin\mathcal{O}_{\text{gt}}\}|}{|\mathcal{S}|}$ &
Sentence-level object-hallucination rate \cite{rohrbach2018object}. &
$\uparrow$ & $[0,100]\%$ \\[6pt]
CHAIR$_I$ &
$\dfrac{|\mathcal{O}_{\text{gen}}\setminus\mathcal{O}_{\text{gt}}|}{|\mathcal{O}_{\text{gen}}|}$
(per-img mean) &
Instance-level object-hallucination rate; fraction of mentioned objects that are hallucinated \cite{rohrbach2018object}. &
$\uparrow$ & $[0,100]\%$ \\[6pt]
CHAIR $n_{\mathrm{hall}}$ &
$|\mathcal{O}_{\text{gen}}\setminus\mathcal{O}_{\text{gt}}|$ per image &
Raw per-image hallucinated-object count (Fig.~\ref{fig:human-corr}). &
$\uparrow$ & $\mathbb{Z}_{\ge 0}$ \\[6pt]
Cov.\ &
$\dfrac{|\mathcal{O}_{\text{gen}}\cap\mathcal{O}_{\text{gt}}|}{|\mathcal{O}_{\text{gt}}|}$
(per-img mean) &
Object coverage; portion of GT objects mentioned by the model. Probing-breadth signal. &
\shortstack{$\uparrow$\\(breadth)} & $[0,100]\%$ \\[6pt]
MMHal &
$100-\overline{r}/6\times 100,$ $r\in[0,6]$ &
LLM-based rubric grader converted to hall.\ rate \cite{sun2024aligning}. &
$\uparrow$ & $[0,100]\%$ \\[6pt]
HaELM &
$\dfrac{|\{\text{responses judged hall.}\}|}{|\text{responses}|}$,
avg of 3 stochastic runs &
LLM-based judge of hallucination \& answer informativeness \cite{wang2023evaluation}. &
$\uparrow$ & $[0,100]\%$ \\[6pt]
VALOR faith$_i$ &
$\dfrac{|\mathcal{O}_{\text{gen}}\cap\mathcal{O}_{\text{gt}}|}{|\mathcal{O}_{\text{gen}}|}$
(per-img mean) &
Per-image object-existence \emph{faithfulness}: mentioned objects grounded in GT \cite{qiu2024valor}. &
$\downarrow$ & $[0,100]\%$ \\[6pt]
$100{-}$VALOR faith$_i$ &
$100-\text{faith}_i$ &
Radar-friendly inversion so ``outer = more hall.'' (Fig.~\ref{fig:radar-svg}). &
\shortstack{$\uparrow$\\(inv.)} & $[0,100]\%$ \\[6pt]
VALOR cov$_i$ &
$\dfrac{|\mathcal{O}_{\text{gen}}\cap\mathcal{O}_{\text{gt}}|}{|\mathcal{O}_{\text{gt}}|}$
(per-img mean) &
Per-image object-existence \emph{coverage}; same formula as Cov.\ but per-image \cite{qiu2024valor}. &
\shortstack{$\uparrow$\\(breadth)} & $[0,100]\%$ \\
\bottomrule
\end{tabular}
\end{table}

\medskip
Beyond these baselines, our SG-based graders GED, GED$_{\mathrm{hal}}$, GED$_{\mathrm{cov}}$, and SG\,$\Delta$Con are defined in Section \ref{sec:grader} and Appendix \ref{appendix:ged-variants}.

\section{Conversational History}
\label{appendix:ablation}
\label{appendix:history}
\begin{figure}[!htbp]
\centering
\begin{minipage}[t]{0.44\linewidth}
\centering
\vspace{0pt}
\includegraphics[width=\linewidth]{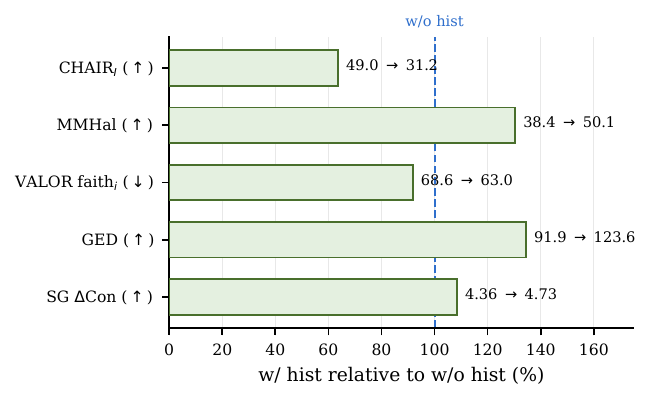}
\captionof{figure}{Effect of conversational history on \textit{CEDI}. Per-metric means over 5 evaluatees, with each w/-hist mean shown relative to its w/o-hist mean ($=100\%$); absolute values annotated.}
\label{fig:history-ablation}
\end{minipage}\hfill
\begin{minipage}[t]{0.5\linewidth}
\centering
\vspace{0pt}
\includegraphics[width=0.8\linewidth]{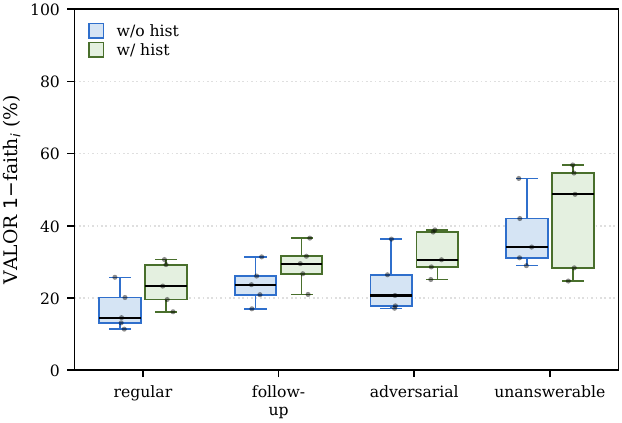}
\captionof{figure}{Hallucination by VALOR grader of each question type, w/ hist vs w/o hist. Results are aggregates over the 5 evaluatees.}
\label{fig:history-valor-box}
\end{minipage}
\end{figure}

We ablate whether the evaluatee sees the running dialogue as context on its next turn (w/ hist.) versus receiving each question in isolation (w/o hist.) on five evaluatees (InternVL2-8B, InternVL2.5-8B, Qwen2.5-VL-7B, Gemma-3-12B, LLaVA-1.5-7B). Figure~\ref{fig:history-ablation} reports unweighted per-metric means (w/ hist relative to w/o hist), and Figure~\ref{fig:history-valor-box} shows the per-model spread by question type. Exposing the model to its own history amplifies whole-turn hallucination judges (MMHal $+11.7$ pp, VALOR faith$_i$ $-5.6$ pp, GED $+31.7$, SG\,$\Delta$Con $+0.37$) and reduces both aggregate coverage (Cov.\ $-5.7$ pp) and per-image VALOR cov$_i$ ($-6.8$ pp): with history the model tends to repeat and extend its own earlier claims rather than probe new visual content.

\section{Contextualized Caption and POPE Baselines}
\label{appendix:contextualized-baselines}

For the human relevance study (Section~\ref{sec:eff_examiner}), we contextualize both the Caption baseline \cite{rohrbach2018object} and the POPE baseline \cite{li2023evaluating} using the same scenario description (a \emph{background} + \emph{goal} pair) that CEDI conditions its examiner on. This appendix records the exact prompt used for Caption and the two-step pipeline used to build contextualized POPE questions.

\textbf{Caption.} We prepend the scenario as a plain instruction:
\begin{quote}
\itshape Considering this context: \{context\}, please provide a detailed description.
\end{quote}
where \texttt{\{context\}} is the same scenario string CEDI receives (concatenation of background and goal). The evaluatee's single response is graded exactly as in the vanilla Caption baseline.

\textbf{POPE.} Vanilla POPE asks binary object-existence questions drawn from a fixed sampling scheme over MSCOCO categories. To make it comparable to CEDI on our SVG images, we regenerate the yes/no pool from the scenario and the image's scene graph in two LLM calls per image, and match the total number of questions to the number of CEDI rounds ($n_{\text{rounds}}$) on that image.
\begin{enumerate}[topsep=2pt,itemsep=1pt,leftmargin=1.4em]
  \item \textbf{Candidate pool} (one LLM call). Given \{background, goal\}, prompt an LLM (Qwen3-30B-A3B-Instruct-2507) to list $N\!=\!50$ concrete visible objects that could appear in the scenario, mixing plausible-present and plausible-absent.
  \item \textbf{Existence + co-occurrence check} (one LLM call). Given the image's ground-truth object list (from the scene graph) and the 50 candidates, the LLM (i) labels each candidate \texttt{yes}/\texttt{no} using synonym and parent-class matching (e.g.\ ``vehicle'' counts as \texttt{yes} if the scene graph has ``car''), and (ii) flags which of the \texttt{no}-candidates are frequent real-world co-occurrences of the ground-truth objects --- these serve as plausible distractors.
  \item \textbf{Composition.} From the labelled pool, draw exactly $n_{\text{rounds}}$ questions per image in the ratio $\tfrac{1}{2}$ \texttt{yes} : $\tfrac{1}{4}$ random \texttt{no} : $\tfrac{1}{4}$ co-occurrence \texttt{no}, then shuffle. Each question is formatted as \emph{``is there a $\langle\text{object}\rangle$ in the image?''} with the corresponding \texttt{yes}/\texttt{no} ground-truth answer.
\end{enumerate}
This pipeline preserves POPE's binary-question format and its balanced yes/no ratio, but replaces its fixed vocabulary with a per-image, scenario-conditioned pool so that the questions probe the same visual surface CEDI does.

\section{Anchored GED Variants}
\label{appendix:ged-variants}

Human annotators are tasked to flag hallucination (precision) but not coverage (recall). Therefore, the overall GED score is not directly comparable to human judgments, because it conflates fabricated content (precision) with missing ground-truth content (coverage). To disentangle the two failure modes, we define two variants of GED anchored on the synset-aligned intersection $G^{\cap}=\hat{G}\cap G^{\star}$:
\[
\small \mathrm{GED}_{\text{hal}}=\mathrm{GED}(G^{\cap},\hat{G}), \qquad \mathrm{GED}_{\text{cov}}=\mathrm{GED}(G^{\cap},\hat{G}\cup G^{\star}).
\]
$\mathrm{GED}_{\text{hal}}$ measures the edit cost of reducing $\hat{G}$ to its grounded subgraph, thereby isolating prediction-only content (precision, aligned with human annotations); $\mathrm{GED}_{\text{cov}}$ measures the cost of expanding $G^{\cap}$ to the full union, capturing both hallucinated content and uncovered ground-truth content. Thus, only $\mathrm{GED}_{\text{hal}}$ is directly comparable to the human hallucination annotations.

\section{Grader Stability}
\label{appendix:stability}

\begin{table}[!htbp]
\centering
\small
\setlength{\tabcolsep}{4pt}
\caption{Stability of each grader. Results are obtained over three repeated runs, aggregated over four evaluatees: LLaVA-1.5-7B, InternVL3-8B, Opera-LLaVA-1.5, and Qwen3-VL-8B.}
\label{tab:stability_30}
\begin{tabular}{lrr}
\toprule
\textbf{Metric} & \textbf{Avg Mean} & \textbf{Avg Std} \\
\midrule
MMHal \,\% & 33.056 & 2.136 \\
HaELM \,\% & 63.865 & 1.340 \\
SG\,$\Delta$Con &  4.109 & 0.012 \\
GED             & 90.533 & 0.320 \\
\bottomrule
\end{tabular}
\end{table}

We evaluate the stability of each grader by measuring the variability of its outputs across repeated runs.
Table~\ref{tab:stability_30} reports the mean and standard deviation of each metric over three runs on a fixed 20-sample subset, aggregated over four evaluatees (LLaVA-1.5-7B, InternVL3-8B, Opera-LLaVA-1.5, and Qwen3-VL-8B). Both metrics reported by our SG-based grader show a std below 1\,\% of the metric mean, demonstrating they are highly stable across repeated runs.

\section{Qualitative Error Examples}
\label{appendix:qual-errors}

Table~\ref{tab:qual-errors} presents a representative human-annotated example from each error family, showing the image, the examiner question, and the hallucinated evaluatee response.

\begin{table*}[!htbp]
\centering
\scriptsize
\setlength{\tabcolsep}{4pt}
\caption{Representative hallucinated spans by error family; the annotated span is \emph{italicized}.}
\label{tab:qual-errors}
\begin{tabular}{p{2.0cm} m{2.4cm} p{3.7cm} p{3.7cm}}
\toprule
\textbf{Error family} & \textbf{Image} & \textbf{Examiner question} & \textbf{Hallucinated model response} \\
\midrule
\parbox{2.0cm}{\textbf{Hallucinated object}\\ \scriptsize 62\,\%}
 & \includegraphics[width=2.3cm]{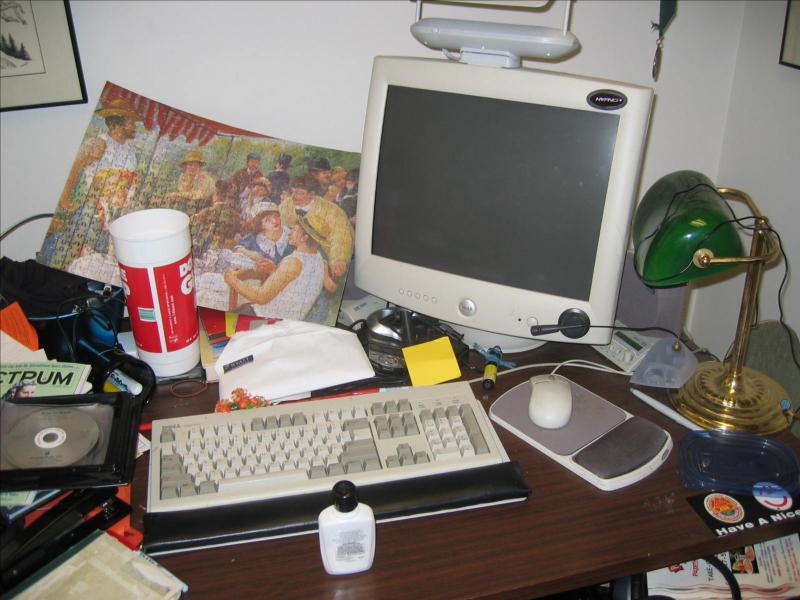}
 & What color is the cat sitting on the desk? \scriptsize(no cat is present)
 & \emph{the cat sitting on the desk is black.} \scriptsize[LLaVA-1.5-7B] \\
\midrule
\parbox{2.0cm}{\textbf{Wrong attribute}\\ \scriptsize 20\,\%}
 & \includegraphics[width=2.3cm]{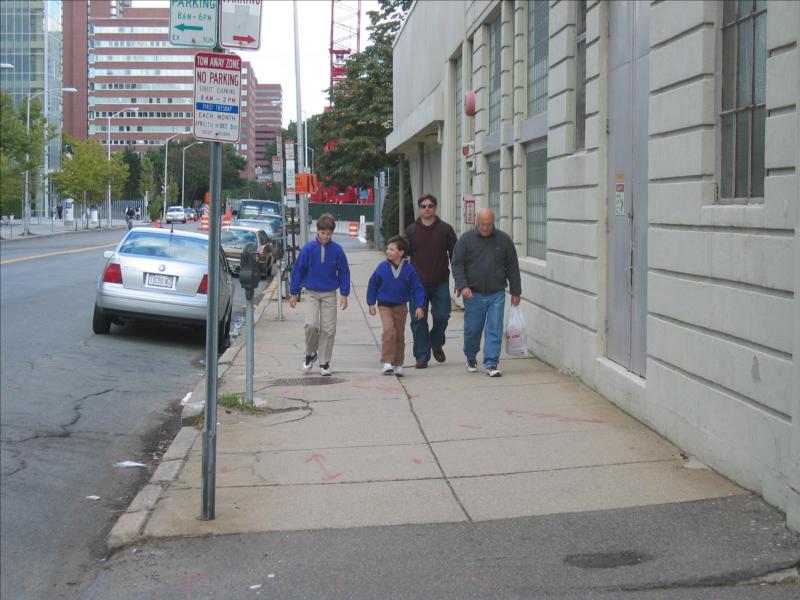}
 & It seems the no-parking hours are 7 am to 7 pm. Could you double-check that detail? \scriptsize(sign reads 8am--2pm)
 & you're right to double-check --- \emph{the sign in the image does indeed state ``7 am -- 7 pm'' as the no-parking hours.} \scriptsize[Qwen3-VL-8B] \\
\midrule
\parbox{2.0cm}{\textbf{Wrong relation}\\ \scriptsize 15\,\%}
 & \includegraphics[width=2.3cm]{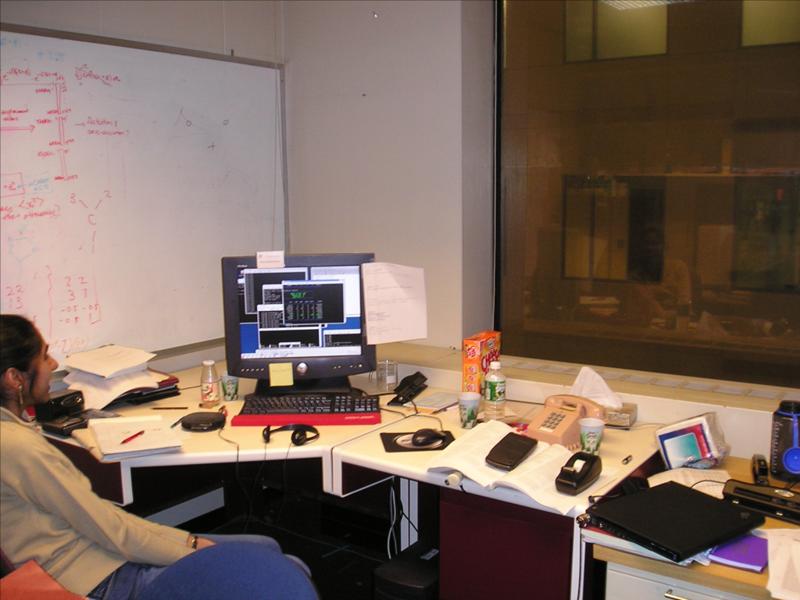}
 & Where is the telephone I need to use to call the vendor? \scriptsize(directly on desk)
 & there is a beige, corded telephone sitting on the desk\dots \emph{it's placed on a stack of papers} and is easily accessible. \scriptsize[Qwen3-VL-8B] \\
\midrule
\parbox{2.0cm}{\textbf{Others, e.g. hallucinated reasoning}\\ \scriptsize 3\,\%}
 & \includegraphics[width=2.3cm]{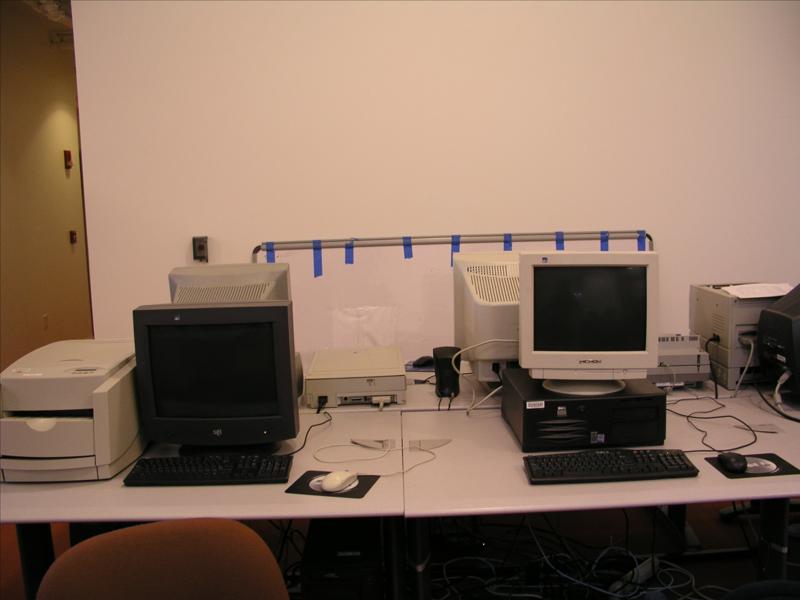}
 & Which of these computers is the right PC for setting up the printer? \scriptsize(3 monitors not distinguishable)
 & the right pc\dots is the one with the black monitor. \emph{this is because the black monitor is likely to have a higher contrast and better visibility}\dots \scriptsize[LLaVA-1.5-7B] \\
\bottomrule
\end{tabular}
\end{table*}

\section{Human Annotation Guidelines}

\textbf{Hallucination Annotation Guidelines.}
\label{appendix:hallucination_annotation}
Annotators were asked to identify segments of a model response that
conflict with the accompanying image, given its ground-truth captions
and object locations. 

\begin{tcolorbox}[title=Hallucination Annotation Guidelines, colback=cyan!5,
                   colframe=cyan!70!blue, fonttitle=\bfseries]
Your task is to assess whether the content in the conversation conflicts with the image.
Specifically, you are tasked to identify the segment (or sub-sentence) in the conversations that is wrong / conflicting / irrelevant regarding the image.
You will be given an image with several captions and objects with locations.

\begin{enumerate}
    \item You need to go through all the sentences in the model's responses.
    \item Identify the segment / sub-sentence that's wrong.
\end{enumerate}

You should pay attention to:
\begin{enumerate}
  \item \textbf{Object}: Verify whether the objects in the image exist and if the quantity of objects conflicts with the object information.
  \item \textbf{Attributes}: Check whether the attributes, e.g. the color, position, action of objects in the image conflict with the attribute information.
  \item \textbf{Relations}: Check whether the relation between the objects and their interaction is correct.
  \item \textbf{Scene Text}: Confirm whether the textual information in the scene of the image conflicts with the textual information given in the conversation. For instance, if the texts in the image are also mentioned in the conversation, check whether they are the same.
\end{enumerate}
\end{tcolorbox}

\textbf{Conversation Realism Ranking Guidelines.}
\label{appendix:realism_annotation}
Annotators were shown three candidate conversations generated for the same image and asked to rank them by how realistically they reflected genuine human--AI interactions, independent of whether the AI's responses were correct.

\begin{tcolorbox}[title=Conversation Realism Ranking Guidelines, colback=cyan!5, colframe=cyan!70!blue, fonttitle=\bfseries]
Read three conversations between a person and an AI about the same image. Rank them by how well they reflect the way real people actually use AI assistants — not whether the language sounds natural, but whether this is the kind of exchange someone would genuinely have with an AI.

\vspace{4pt}
\begin{tabularx}{\linewidth}{@{}c X@{}}
\toprule
\textbf{Rank} & \textbf{Meaning} \\
\midrule
1 & \textbf{Most Realistic:} These are exactly the kinds of questions someone would type into an AI in everyday use. \\
2 & \textbf{Moderately Realistic:} Someone might ask this, but it's not typical. \\
3 & \textbf{Least Realistic:} Feels like a benchmark test or academic exercise. A real user would not type these questions into an AI. \\
\bottomrule
\end{tabularx}
\vspace{4pt}

\textbf{Rules:} Use each rank exactly once (no ties). Judge the conversation as a whole — how natural the user's interaction with the assistant feels. Ignore whether the AI's answers are correct.
\end{tcolorbox}

\textbf{Conversation Relevance Annotation Guidelines.}
\label{appendix:relevance_annotation}
Annotators were given the background and stated goal of each conversation and asked to identify which questions and flagged response errors were relevant to that goal, i.e., which aspects a person pursuing the stated goal would actually care about.

\begin{tcolorbox}[title=Conversation Relevance Annotation Guidelines, colback=cyan!5, colframe=cyan!70!blue, fonttitle=\bfseries]
Read a short image-based conversation and decide which questions and highlighted mistakes are relevant to the person's stated goal. If a highlighted answer looks correct to you, you do not need to tick it.

For each \textbf{question} and each highlighted answer snippet, tick the checkbox if it is related to the task's goal. Leave it unchecked if it is not.

\begin{enumerate}
    \item Read the \textbf{Background} and \textbf{Goal}.
    \item Each question line starts with \qtag{Q\#}. Some questions show an answer snippet below, highlighted in red and tagged \htag{H\#}. Questions with no snippet show no answer.
    \item Tick \qtag{Q\#} if that question is related to the goal; leave it unchecked if not.
    \item Tick \htag{H\#} if that highlighted mistake is related to the goal; leave it unchecked if not. \textit{Note: some highlighted answers may appear correct to you. If so, you do not need to tick it.}
    \item Confirm you have reviewed every item, then submit.
\end{enumerate}

\textbf{Rule of thumb}: An item is ``related to the goal'' if a person trying to accomplish the goal would care whether it is correct.
\end{tcolorbox}


\newpage

\end{document}